\documentclass[11pt,a4paper]{article}
\usepackage[hyperref]{emnlp2018}
\usepackage{times}
\usepackage{latexsym}
\usepackage{booktabs}% for better rules in the table
\usepackage{multirow}
\usepackage{makecell}
\usepackage{url}
\usepackage{times}
\usepackage{latexsym}
\usepackage{amsmath,amssymb}%,amsthm,mathtools}
\usepackage{graphicx}
\usepackage{subcaption}
\usepackage{titlesec}
\usepackage{comment}
\usepackage{tabularx}

\aclfinalcopy 

\title{Reducing Gender Bias in Abusive Language Detection}

\author{Ji Ho Park, Jamin Shin, Pascale Fung \\
  Centre for Artificial Intelligence Research (CAiRE) \\
  Hong Kong University of Science and Technology \\
  {\tt \{jhpark, jmshinaa\}@connect.ust.hk},  {\tt pascale@ece.ust.hk}} 

\date{}

\begin{document}
\maketitle
\begin{abstract}
Abusive language detection models tend to have a problem of being biased toward identity words of a certain group of people because of imbalanced training datasets. For example, ``You are a good \textit{woman}'' was considered ``sexist'' when trained on an existing dataset. Such model bias is an obstacle for models to be robust enough for practical use. In this work, we measure gender biases on models trained with different abusive language datasets, while analyzing the effect of different pre-trained word embeddings and model architectures. We also experiment with three bias mitigation methods: (1) debiased word embeddings, (2) gender swap data augmentation, and (3) fine-tuning with a larger corpus. These methods can effectively reduce gender bias by 90-98\% and can be extended to correct model bias in other scenarios. 
\end{abstract}

\section{Introduction}
Automatic detection of abusive language is an important task since such language in online space can lead to personal trauma, cyber-bullying, hate crime, and discrimination. As more and more people freely express their opinions in social media, the amount of textual contents produced every day grows almost exponentially, rendering it difficult to effectively moderate user content. For this reason, using machine learning and natural language processing (NLP) systems to automatically detect abusive language is useful for many websites or social media services.

Although many works already tackled on training machine learning models to automatically detect abusive language, recent works have raised concerns about the robustness of those systems. \citet{hosseini2017deceiving} have shown how to easily cause false predictions with adversarial examples in Google's API, and \citet{dixon2017measuring} show that classifiers can have unfair biases toward certain groups of people.

We focus on the fact that the representations of abusive language learned in only supervised learning setting may not be able to generalize well enough for practical use since they tend to overfit to certain words that are neutral but occur frequently in the training samples. To such classifiers, sentences like ``You are a good woman'' are considered ``sexist'' probably because of the word ``woman.'' 

This phenomenon, called \textit{false positive bias}, has been reported by \citet{dixon2017measuring}. They further defined this model bias as unintended, ``a model contains unintended bias if it performs better for comments containing some particular identity terms than for comments containing others.'' 

Such model bias is important but often unmeasurable in the usual experiment settings since the validation/test sets we use for evaluation are already biased. For this reason, we tackle the issue of measuring and mitigating unintended bias. Without achieving certain level of generalization ability, abusive language detection models may not be suitable for real-life situations.

In this work, we address model biases specific to gender identities (gender bias) existing in abusive language datasets by measuring them with a generated unbiased test set and propose three reduction methods: (1) debiased word embedding, (2) gender swap data augmentation, (3) fine-tuning with a larger corpus. Moreover, we compare the effects of different pre-trained word embeddings and model architectures on gender bias.

\section{Related Work}
So far, many efforts were put into defining and constructing abusive language datasets from different sources and labeling them through crowd-sourcing or user moderation \cite{waseem2016hateful,waseem2016you,founta2018large,wulczyn2017ex}. Many deep learning approaches have been explored to train a classifier with those datasets to develop an automatic abusive language detection system \cite{badjatiya2017deep,park2017one,pavlopoulos2017deeper}. However, these works do not explicitly address any model bias in their models.

Addressing biases in NLP models/systems have recently started to gain more interest in the research community, not only because fairness in AI is important but also because bias correction can improve the robustness of the models. \citet{bolukbasi2016man} is one of the first works to point out the gender stereotypes inside word2vec \cite{mikolov2013distributed} and propose an algorithm to correct them. \citet{caliskan2017semantics} also propose a method called Word Embedding Association Test (WEAT) to measure model bias inside word embeddings and finds that many of those pretrained embeddings contain problematic bias toward gender or race. \citet{dixon2017measuring} is one of the first works that point out existing ``unintended'' bias in abusive language detection models. \citet{kiritchenko2018examining} compare 219 sentiment analysis systems participating in SemEval competition with their proposed dataset, which can be used for evaluating racial and gender bias of those systems. \citet{zhao2018gender} shows the effectiveness of measuring and correcting gender biases in co-reference resolution tasks. We later show how we extend a few of these works into ours.

\section{Datasets}
\label{sec:datasets}

\subsection{Sexist Tweets (\texttt{st})}
\label{sec:waseem}
This dataset consists of tweets with sexist tweets collected from Twitter by searching for tweets that contain common terms pertaining to sexism such as ``feminazi.'' The tweets were then annotated by experts based on criteria founded in critical race theory. The original dataset also contained a relatively small number of ``racist'' label tweets, but we only retain ``sexist'' samples to focus on gender biases. \citet{waseem2016hateful,waseem2016you}, the creators of the dataset, describe ``sexist'' and ``racist'' languages as specific subsets of abusive language. 

\subsection{Abusive Tweets (\texttt{abt})}
\label{sec:founta}
Recently, \citet{founta2018large} has published a large scale crowdsourced abusive tweet dataset with 60K tweets. Their work incrementally and iteratively investigated methods such as boosted sampling and exploratory rounds, to effectively annotate tweets through crowdsourcing. Through such systematic processes, they identify the most relevant label set in identifying abusive behaviors in Twitter as $\{None, Spam, Abusive, Hateful\}$ resulting in 11\% as 'Abusive,' 7.5\% as 'Hateful', 22.5\% as 'Spam', and 59\% as 'None'. We transform this dataset for a binary classification problem by concatenating 'None'/'Spam' together, and 'Abusive'/'Hateful' together.

\begin{table}[t] 
\footnotesize
	\centering
\begin{tabular}{@{}cccccc@{}}
	\toprule
Name & Size & Positives (\%) & $\mu$ & $\sigma$ & $max$ \\
	\midrule
\texttt{st} & 18K & 33\% & 15.6 & 6.8 & 39 \\ 
\texttt{abt} & 60K & 18.5\% & 17.9 & 4.6 & 65 \\ 
	\bottomrule
\end{tabular}
\caption{Dataset statistics. $\mu, \sigma, max$ are mean, std.dev, and maximum of sentence lengths} \label{table:dataset_stats}
\end{table}

\section{Measuring Gender Biases}
\subsection{Methodology}
Gender bias cannot be measured when evaluated on the original dataset as the test sets will follow the same biased distribution, so normal evaluation set will not suffice. Therefore, we generate a separate \textit{unbiased test set} for each gender, male and female, using the identity term template method proposed in \citet{dixon2017measuring}. 

The intuition of this template method is that given a pair of sentences with only the identity terms different (ex. ``He is happy'' \& ``She is happy''), the model should be able to generalize well and output same prediction for abusive language. This kind of evaluation has also been performed in \textit{SemEval 2018: Task 1 Affect In Tweets} \cite{kiritchenko2018examining} to measure the gender and race bias among the competing systems for sentiment/emotion analysis. 

Using the released code\footnote{\url{https://github.com/conversationai/unintended-ml-bias-analysis}} of \citet{dixon2017measuring}, we generated 1,152 samples (576 pairs) by filling the templates with common gender identity pairs (ex. male/female, man/woman, etc.). We created templates (Table \ref{table:templates}) that contained both neutral and offensive nouns and adjectives inside the vocabulary (See Table \ref{table:offensive-words}) to retain balance in neutral and abusive samples.

\begin{table}[t]
\footnotesize
\begin{center}
\begin{tabular}{|c|}
\hline \bf{Example Templates} \\ 
\hline
You are a \textit{(adjective) (identity term)}. \\
\textit{(verb) (identity term)}. \\
Being \textit{(identity term)} is \textit{(adjective)} \\
I am \textit{(identity term)} \\
I hate \textit{(identity term)} \\
\hline
\end{tabular}
\end{center}
\caption{\label{table:templates} Example of templates used to generated an unbiased test set.}
\end{table}

\begin{table}[t]
\footnotesize
\begin{center}
\begin{tabularx}{\linewidth}{|X|X|}
\hline \bf{Type} & \bf{Example Words} \\ 
\hline
Offensive & disgusting, filthy, nasty, rude, horrible, terrible, awful, worst, idiotic, stupid, dumb, ugly, etc. \\
\hline
Non-offensive & help, love, respect, believe, congrats, hi, like, great, fun, nice, neat, happy, good, best, etc. \\
\hline
\end{tabularx}
\end{center}
\caption{\label{table:offensive-words} Example of offensive and non-offensive verbs \& adjectives used for generating the unbiased test set.}
\end{table}

For the evaluation metric, we use 1) AUC scores on the original test set (Orig. AUC), 2) AUC scores on the unbiased generated test set (Gen. AUC), and 3) the false positive/negative equality differences proposed in \citet{dixon2017measuring} which aggregates the difference between the overall false positive/negative rate and gender-specific false positive/negative rate. False Positive Equality Difference (FPED) and False Negative Equality Difference (FNED) are defined as below, where $T = \{male,female\}$. 
\begin{gather*} 
\footnotesize
FPED = \sum_{t\in T} | FPR - FPR_t| \\
FNED = \sum_{t\in T} | FNR - FNR_t| 
\end{gather*} 
Since the classifiers output probabilities, equal error rate thresholds are used for prediction decision. 

While the two AUC scores show the performances of the models in terms of accuracy, the equality difference scores show them in terms of fairness, which we believe is another dimension for evaluating the model's generalization ability.

\subsection{Experimental Setup}
\label{sec: experiment}
We first measure gender biases in \texttt{st} and \texttt{abt} datasets. We explore three neural models used in previous works on abusive language classification: Convolutional Neural Network (CNN) \cite{park2017one}, Gated Recurrent Unit (GRU) \cite{cho2014learning}, and Bidirectional GRU with self-attention ($\alpha$-GRU) \cite{pavlopoulos2017deeper}, but with a simpler mechanism used in \citet{felbo2017using}. Hyperparameters are found using the validation set by finding the best performing ones in terms of original AUC scores. These are the used hyperparameters:

\begin{enumerate}
\item CNN: Convolution layers with 3 filters with the size of [3,4,5], feature map size=100, Embedding Size=300, Max-pooling, Dropout=0.5
\item GRU: hidden dimension=512, Maximum Sequence Length=100, Embedding Size=300, Dropout=0.3
\item $\alpha$-GRU: hidden dimension=256 (bidirectional, so 512 in total), Maximum Sequence Length=100, Attention Size=512, Embedding Size=300, Dropout=0.3
\end{enumerate}

We also compare different pre-trained embeddings, \texttt{word2vec} \cite{mikolov2013distributed} trained on Google News corpus, \texttt{FastText} \cite{bojanowski2016enriching}) trained on Wikipedia corpus, and randomly initialized embeddings (\textit{random}) to analyze their effects on the biases. Experiments were run 10 times and averaged.

\begin{table}[t]
\footnotesize
\begin{center}
\scalebox{0.9}{
\begin{tabular}{|l|l|c|c|c|c|}
\hline
\thead{\bf{Model}} & \thead{\bf{Embed.}} & \thead{\bf{Orig.}\\ \bf{AUC}}  & \thead{\bf{Gen.} \\ \bf{AUC}} & \thead{\bf{FNED}} & \thead{\bf{FPED}} \\
\hline \multirow{3}{*}{{CNN}}          & random & .881        & .572      & .261      & .249 \\
                                       & fasttext & .\bf{906} & .620      & .323      & .327 \\
                                       & word2vec & \bf{.906} & .635      & .305      & .263  \\
\hline \multirow{3}{*}{{GRU}}          & random & .854        & .536      & \bf{.132} & .\bf{136} \\
                                       & fasttext & .887      & \bf{.661} & .312      & .284 \\
                                       & word2vec & .887      & .633      & .301      & .254  \\
\hline \multirow{3}{*}{{$\alpha$-GRU}} & random   & .868      & .586      & .236      & .219 \\
                                       & fasttext & .891      & .639      & .324      & .365 \\
                                       & word2vec & .890      & .631      & .315      & .306  \\
\hline
\end{tabular}
}
\end{center}
\caption{\label{table:waseem-result} Results on \texttt{st}. False negative/positive equality differences are larger when pre-trained embedding is used and CNN or $\alpha$-RNN is trained}
\end{table}

\subsection{Results \& Discussions}
\label{sec:measurement-results}
Tables \ref{table:waseem-result} and \ref{table:founta-result} show the bias measurement experiment results for \texttt{st} and \texttt{abt}, respectively. As expected, pre-trained embeddings improved task performance. The score on the unbiased generated test set (Gen. ROC) also improved since word embeddings can provide prior knowledge of words.

However, the equality difference scores tended to be larger when pre-trained embeddings were used, especially in the \texttt{st} dataset. This confirms the result of \citet{bolukbasi2016man}. In all experiments, direction of the gender bias was towards female identity words. We can infer that this is due to the more frequent appearances of female identities in ``sexist'' tweets and lack of negative samples, similar to the reports of \citet{dixon2017measuring}. This is problematic since not many NLP datasets are large enough to reflect the true data distribution, more prominent in tasks like abusive language where data collection and annotation are difficult.

On the other hand, \texttt{abt} dataset showed significantly better results on the two equality difference scores, of at most 0.04. Performance in the generated test set was better because the models successfully classify abusive samples regardless of the gender identity terms used. Hence, we can assume that \texttt{abt} dataset is less gender-biased than the \texttt{st} dataset, presumably due to its larger size, balance in classes, and systematic collection method. 

Interestingly, the architecture of the models also influenced the biases. Models that ``attend'' to certain words, such as CNN's max-pooling or $\alpha$-GRU's self-attention, tended to result in higher false positive equality difference scores in \texttt{st} dataset. These models show effectiveness in catching not only the discriminative features for classification, but also the ``unintended'' ones causing the model biases.

\begin{table}[t]
\footnotesize
\begin{center}
\scalebox{0.9}{
\begin{tabular}{|l|l|c|c|c|c|}
\hline
\thead{\bf{Model}} & \thead{\bf{Embed.}} & \thead{\bf{Orig.}\\ \bf{AUC}}  & \thead{\bf{Gen.} \\ \bf{AUC}} & \thead{\bf{FNED}} & \thead{\bf{FPED}} \\
\hline \multirow{3}{*}{{CNN}} & random & .926        & .893      & .013      & .045 \\
                              & fasttext & .955      & .995      & .004      & \bf{.001} \\
                              & word2vec & \bf{.956} & \bf{.999} & \bf{.002} & .021  \\
\hline \multirow{3}{*}{{GRU}} & random & .919        & .850      & .036      & .010 \\
                              & fasttext & .951      & .997      & .014      & .018 \\
                              & word2vec & .952      & .997      & .017      & .037  \\
\hline \multirow{3}{*}{{$\alpha$-GRU}} & random & .927      & .914      & .008      & .039 \\
                              & fasttext & \bf{.956} & .998      & .014      & .005 \\
                              & word2vec & .955      & \bf{.999} & .012      & .026  \\
\hline
\end{tabular}
}
\end{center}
\caption{\label{table:founta-result} Results on \texttt{abt}. The false negative/positive equality difference is significantly smaller than the \texttt{st}}
\end{table}

\section{Reducing Gender Biases}
We experiment and discuss various methods to reduce gender biases identified in Section \ref{sec:measurement-results}.
\subsection{Methodology}

\noindent \textbf{Debiased Word Embeddings (DE)} \cite{bolukbasi2016man} proposed an algorithm to correct word embeddings by removing gender stereotypical information. All the other experiments used pretrained word2vec to initialized the embedding layer but we substitute the pretrained word2vec with their published embeddings to verify their effectiveness in our task. 

\noindent \textbf{Gender Swap (GS)} We augment the training data by identifying male entities and swapping them with equivalent female entities and vice-versa. This simple method removes correlation between gender and classification decision and has proven to be effective for correcting gender biases in co-reference resolution task \cite{zhao2018gender}.

\noindent \textbf{Bias fine-tuning (FT)} We propose a method to use transfer learning from a less biased corpus to reduce the bias. A model is initially trained with a larger, less-biased source corpus with a same or similar task, and fine-tuned with a target corpus with a larger bias. This method is inspired by the fact that model bias mainly rises from the imbalance of labels and the limited size of data samples. Training the model with a larger and less biased dataset may regularize and prevent the model from over-fitting to the small, biased dataset.

\begin{table}[t]
\footnotesize
\begin{center}
\scalebox{0.85}{
\begin{tabular}{|c|ccc|c|c|c|c|}
\hline
\thead{\bf{Model}} & \thead{\bf{DE}} & \thead{\bf{GS}} & \thead{\bf{FT}} & \thead{\bf{Orig.}\\ \bf{AUC}}  & \thead{\bf{Gen.} \\ \bf{AUC}} & \thead{\bf{FNED}} & \thead{\bf{FPED}} \\
\hline \multirow{6}{*}{{CNN}} &  . & . & . &  \bf{\underline{.906}} & .635      & .305      & .263 \\
                               & O & . & .                 &  .902 & .627      & .333      & .337 \\
                               & . & O & .                 &  .898 & .676      & .164      & .104 \\
                               & O & O & .                 &  .895 & .647      & .157      & .096 \\                     
                               & . & . & O                 &  .896 & .650      & .302      & .240 \\
                               & . & O & O                 &  .889 & .671      & .163      & .122 \\
                               & O & O & O                 &  .884 & \underline{.703}      & \underline{.135} & \underline{.095} \\
\hline \multirow{6}{*}{{GRU}} &  . & . & .      &  \underline{.887} & .633      & .301      & .254 \\
                               & O & . & .                 &  .882 & .658      & .274      & .270 \\
                               & . & O & .                 &  .879 & .657      & .044      & .040 \\
                               & O & O & .                 &  .873 & .667      & \bf{\underline{.006}}      & \bf{\underline{.027}} \\                     
							   & . & . & O                 &  .874 & .761      & .241      & .181 \\
                               & . & O & O                 &  .862 & .768      & .141      & .095 \\                     
                               & O & O & O                 &  .854 & \underline{.854}      & .081 & .059 \\
\hline \multirow{6}{*}{{$\alpha$-GRU}}&  . & . & .& \underline{.890} & .631      & .315       & .306 \\
                               & O & . & .                 &  .885 & .656      & .291      & .330 \\
                               & . & O & .                 &  .879 & .667      & .114      & .098 \\
                               & O & O & .                 &  .877 & .689      & .067      & .059 \\                     
                               & . & . & O                 &  .874 & .756      & .310      & .212 \\
                               & . & O & O                 &  .866 & .814      & .185      & .065  \\                     
                               & O & O & O                 &  .855 & \bf{\underline{.912}}      & \underline{.055}      & \underline{.030} \\
\hline
\end{tabular}
}
\end{center}
\caption{\label{table:mitigate-result} Results of bias mitigation methods on \texttt{st} dataset. `O' indicates that the corresponding method is applied. See Section \ref{sec:analysis} for more analysis.}
\end{table}

\subsection{Experimental Setup}
\textit{Debiased word2vec} \citet{bolukbasi2016man} is compared with the original \textit{word2vec} \cite{mikolov2013distributed} for evaluation. For gender swapping data augmentation, we use pairs identified through crowd-sourcing by \citet{zhao2018gender}.

After identifying the degree of gender bias of each dataset, we select a source with less bias and a target with more bias. Vocabulary is extracted from training split of both sets. The model is first trained by the source dataset. We then remove final softmax layer and attach a new one initialized for training the target. The target is trained with a slower learning rate. Early stopping is decided by the valid set of the respective dataset.

Based on this criterion and results from Section \ref{sec:measurement-results}, we choose the \texttt{abt} dataset as source and \texttt{st} dataset as target for bias fine-tuning experiments. 

\subsection{Results \& Discussion}
\label{sec:analysis}
Table \ref{table:mitigate-result} shows the results of experiments using the three methods proposed. The first rows are the baselines without any method applied. We can see from the second rows of each section that debiased word embeddings alone do not effectively correct the bias of the whole system that well, while gender swapping significantly reduced both the equality difference scores. Meanwhile, fine-tuning bias with a larger, less biased source dataset helped to decrease the equality difference scores and greatly improve the AUC scores from the generated unbiased test set. The latter improvement shows that the model significantly reduced errors on the unbiased set in general.

To our surprise, the most effective method was applying both debiased embedding and gender swap to GRU, which reduced the equality differences by 98\% \& 89\% while losing only 1.5\% of the original performance. We assume that this may be related to the influence of ``attending'' model architectures on biases as discussed in Section \ref{sec:measurement-results}. On the other hand, using the three methods together improved both generated unbiased set performance and equality differences, but had the largest decrease in the original performance.

All methods involved some performance loss when gender biases were reduced. Especially, fine-tuning had the largest decrease in original test set performance. This could be attributed to the difference in the source and target tasks (abusive \& sexist). However, the decrease was marginal (less than 4\%), while the drop in bias was significant. We assume the performance loss happens because mitigation methods modify the data or the model in a way that sometimes deters the models from discriminating important ``unbiased'' features. 

\section{Conclusion \& Future Work}
We discussed model biases, especially toward gender identity terms, in abusive language detection. We found out that pre-trained word embeddings, model architecture, and different datasets all can have influence. Also, we found our proposed methods can reduce gender biases up to 90-98\%, improving the robustness of the models. 

As shown in Section \ref{sec:measurement-results}, some classification performance drop happens when mitigation methods. We believe that a meaningful extension of our work can be developing bias mitigation methods that maintain (or even increase) the classification performance and reduce the bias at the same time. Some previous works \cite{beutel2017data,zhang2018mitigating} employ adversarial training methods to make the classifiers unbiased toward certain variables. However, those works do not deal with natural language where features like gender and race are latent variables inside the language. Although those approaches are not directly comparable to our methods, it would be interesting to explore adversarial training to tackle this problem in the future.

Although our work is preliminary, we hope that our work can further develop the discussion of evaluating NLP systems in different directions, not merely focusing on performance metrics like accuracy or AUC. The idea of improving models by measuring and correcting gender bias is still unfamiliar but we argue that they can be crucial in building systems that are not only ethical but also practical. Although this work focuses on gender terms, the methods we proposed can easily be extended to other identity problems like racial and to different tasks like sentiment analysis by following similar steps, and we hope to work on this in the future.

\newpage

\section*{Acknowledgments}

This work is partially funded by ITS/319/16FP of Innovation Technology Commission, HKUST, and 16248016 of Hong Kong Research Grants Council.

\bibliography{emnlp2018}
\bibliographystyle{acl_natbib_nourl}

\end{document}